\documentclass{article} 
\usepackage{nips13submit_e,times}
\usepackage{graphicx}
\usepackage{hyperref}
\usepackage{amsfonts}
\usepackage{amsmath}
\usepackage{chngcntr}
\usepackage{url}

\counterwithout{figure}{section}

\title{Sparse, complex-valued representations of natural sounds learned with phase and amplitude continuity priors}

\author{
Wiktor M\l ynarski\\
Max-Planck Institute for Mathematics in the Sciences\\
Leipzig, Germany\\
\texttt{mlynar@mis.mpg.de} \\
}

\nipsfinalcopy 

\begin{document}

\maketitle

\begin{abstract}
Complex-valued sparse coding is a data representation which employs a dictionary of two-dimensional subspaces, while imposing sparse, factorial prior on complex amplitudes. When trained on a dataset of natural image patches, it learns phase invariant features which closely resemble receptive fields of complex cells in the visual cortex. Features trained on natural sounds however, rarely reveal phase invariance and capture other aspects of the data. This observation is a starting point for the present work. As its first contribution, it provides an analysis of natural sound statistics by means of learning sparse, complex representations of short speech intervals. Secondly, it proposes priors over the basis function set, which bias them towards phase-invariant solutions. In this way, a dictionary of complex basis functions can be learned from the data statistics, while preserving the phase invariance property. Finally, representations trained on speech sounds with and without priors are compared. Prior-based basis functions reveal performance comparable to unconstrained sparse coding, while explicitely representing phase as a temporal shift.  Such representations can find applications in many perceptual and machine learning tasks.
\end{abstract}

\section{Introduction}

Natural sounds such as speech, environmental noises or animal calls possess a complex statistical structure. To achieve a good performance in real-world hearing tasks, neuronal and artificial systems should utilize data representations which are adapted to regularities of the natural auditory environment, while making task relevant quantities explicit. 

One particular statistical property of natural sounds, is \emph{sparsness} \cite{Lewicki, SmithLewicki, BellSejnowski}. A typical sample of a sparse signal can be represented as a linear combination of only a few features belonging to a large dictionary. It has been suggested, that neural representations of sound, may take advantage of this regularity. Indeed - learning sparse codes of natural sounds has successfully predicted shapes of cochlear filters \cite{SmithLewicki} and spectrotemporal receptive fields of the Inferior Colliculus neurons \cite{Carlson}, suggesting adaptation of those systems to the natural environment. Sparse coding methods have also proven to be useful in machine hearing applications such as speech and music classification \cite{Grosse}.

However, in some cases linear sparse codes are not an appropriate representation of the data. Linear coding models are not robust to temporal jitter, since even a small temporal stimulus shift changes the encoding \cite{SmithLewickiNC}. Additionally, phase (i.e. temporal displacement not larger than the waveform period) is not explicitly represented. Phase information is relevant for perceptual tasks such as spatial hearing \cite{SmithChimeras} or recognition of conspecific songs in songbirds \cite{Hsu}. For performing such computations, a data representation different from linear sparse coding may be more useful. 

Phase and amplitude information can be separated by representing the data using a combination of complex vectors. Complex-valued sparse coding \cite{CadieuOlshausen} and Independent Component Analysis (ICA) \cite{Laparra} have been applied to learn representations of natural image patches. Real and imaginary parts of complex features emerging from natural images resemble Gabor filters of equal frequency, position scale and orientation in quadrature phase relationship. Such basis functions reveal phase invariance i.e. the value of the complex amplitude does not change with spatial stimulus shifts smaller than the oscillation period. Suggested by results obtained on natural images, one could think that complex sparse codes learned from natural sounds will also reveal phase invariance, since real-valued sparse codes yield features localized in time and frequency \cite{Lewicki, SmithLewicki}. This is, however, not true. Natural sounds possess highly non-local cross-frequency correlations \cite{Terashima}. Reflecting this structure, sparse, complex codes of natural acoustic stimuli capture frequency and bandwidth invariances. Only a small fraction is phase-shift (or time) invariant \cite{WangOlshausenMing}.

In order to learn from the statistics of the data a representation that preserves a desired property such as phase invariance, one could select a parametric form of basis functions and adapt the parameter set \cite{Tosic}. This method has been applied before to audio data by adapting a gammatone dictionary \cite{Yaghoobi}. Despite many advantages of this solution, there exists a possibility, that parametric form of dictionary elements is not flexible enough to efficiently span the data space. To alleviate this problem this paper proposes to learn a sparse and complex representation of natural sounds with the phase-invariance promoting priors. Proposed priors induce temporal continuity, i.e. \emph{slowness} \cite{Foldiak, Wiskott} of both phase and amplitude, which turns out to be a correct assumption for learning phase-invariant features.\\
The goal of the present work is, therefore, twofold:
\begin{enumerate}
\item To analyse high-order statistics of natural sounds. This is done by learning sparse, complex representations of a speech corpus and analysis of the obtained features. Learned dictionaries differ in cardinality (complete and two-times overcomplete) and are learned with and without prior knowledge.
\item To introduce priors useful in learning structured, phase-invariant dictionaries of any data, not only natural sounds. In addition to imposing a desired structure, priors improve convergence of learning algorithms and allow to use less data for learning.
\end{enumerate}

The paper is structured as follows. In section \ref{sec2} complex valued sparse coding model is introduced together with phase and amplitude continuity priors. Section \ref{section3} discusses statistical properties of complex features learned via complete and overcomplete sparse coding on natural, speech sounds. In section \ref{section4} coding efficiency of learned representations is assessed by comparing their performance in a denoising task and estimating entropies of learned coefficients. 

\section{Sparse, complex-valued representations of natural sounds}
\label{sec2}
In a sparse coding model utilizing complex basis functions each data vector $x \in \mathbb{R}^T$ is represented as:
\begin{equation}
\label{eq1}
\hat{x}(t)=\sum_{i=1}^{n/2} \mathfrak{R}\{s_i^*A_i(t)\} + \eta
\end{equation} 
where $s_i \in \mathbb{C}$ are complex coefficients, $*$ denotes a complex conjugation, $A_i \in \mathbb{C}^T$ are complex basis functions and $\eta \sim \mathcal{N}(0, \sigma)$ is an additive gaussian noise.
Complex coefficients in Euler's form become $s_i=a_i e^{j\phi_i}$ (where $j=\sqrt{-1}$) therefore equation \eqref{eq1} can be rewritten to explicitely represent phase $\phi$ and amplitude $a$ as separate variables:
\begin{equation}
\label{eq2}
\hat{x}(t)=\sum_{i=1}^{n/2} a_i\Big(\cos\phi_i A_i^{\mathfrak{R}}(t) + \sin\phi_i A_i^{\mathfrak{I}}(t)\Big) + \eta
\end{equation}
Real and imaginary parts $A_i^{\mathfrak{R}}$ and $A_i^{\mathfrak{I}}$ of basis functions $\{A_i\}_{i=1}^{n/2}$ - span a subspace within which the position of a data sample is determined by amplitude $a_i$ and phase $\phi_i$. Depending on a number of basis functions $n$, the representation can be complete ($n = T$) or overcomplete ($n > T$). If data vectors are chunks of a temporal signal, such as natural sounds, the basis functions $A$ are functions of time (as opposed to basis functions of natural images which constitute a spatial representation). If one attempts to learn a representation of particular time-dependent properties, constraints should be placed on the basis function set. 

\begin{figure}[ht]
\label{fig1}
\begin{center}
\includegraphics{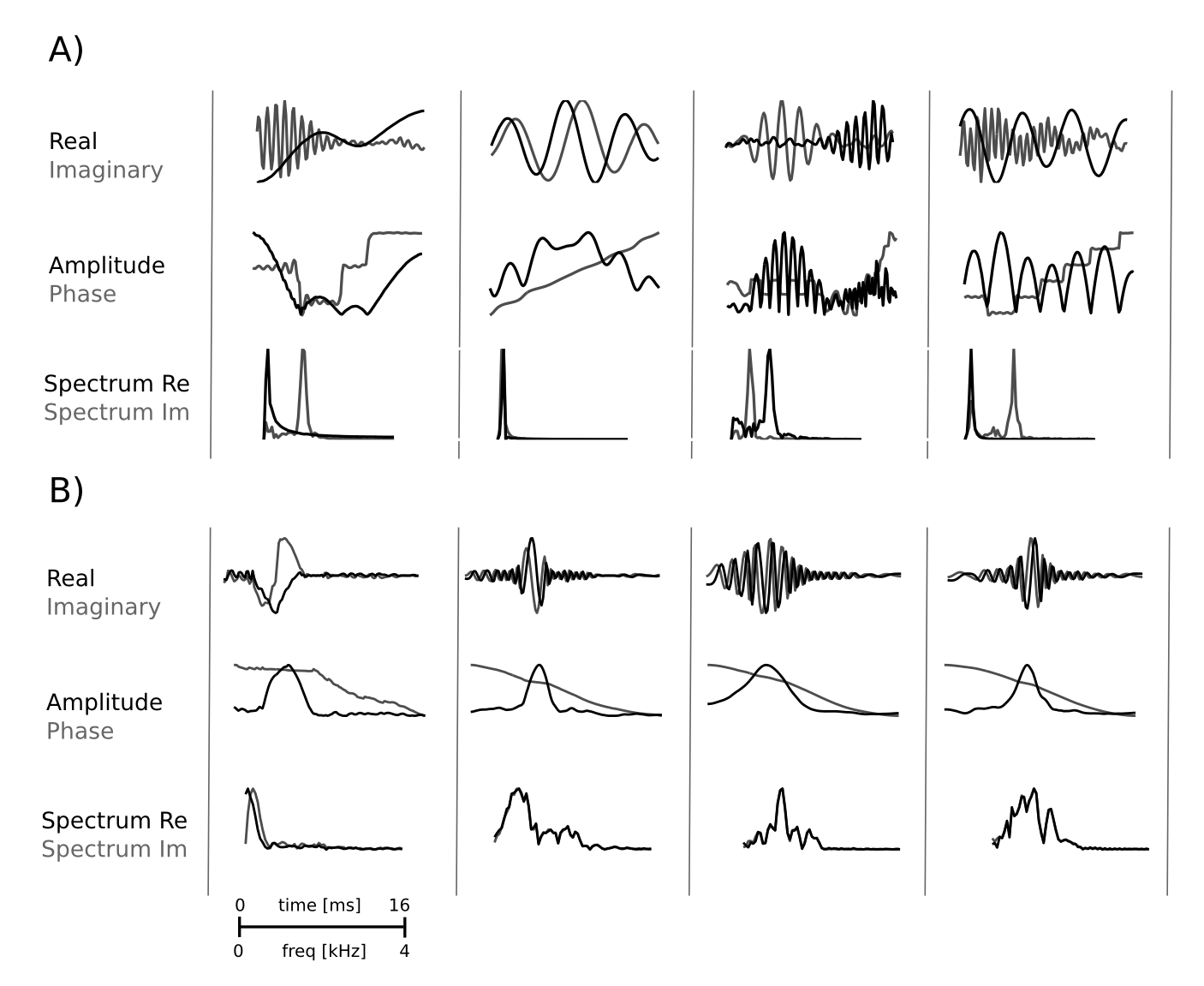}
\end{center}
\caption{Exemplary basis functions learned A) - with and B) - without phase and amplitude continuity priors. First row - basis functions in the time domain, second row - polar coordinates of basis functions, third row - Fourier spectra.}
\end{figure}

In a probabilistic formulation, equations \eqref{eq1} \eqref{eq2} can be understood as a likelihood model of the data, given coefficients $s$ and basis functions $A$:
\begin{equation}
\label{likelihood}
p(x|s,A)=\frac{1}{\sigma \sqrt{2\pi}}\prod_{t=1}^T e^{-\frac{(x(t)-\hat{x}(t))^2}{2\sigma^2}}
\end{equation}
The prior over complex coefficients assumes independence between subspaces and promotes sparse solutions i.e. solutions with most amplitudes close to $0$:
\begin{equation}
\label{coeffInd}
p(s)=\prod_{i=1}^n \frac{1}{Z}e^{-S(a_i)}
\end{equation}
where $Z$ is a normalizing constant. Function  $S(a_i)$ penalizes large amplitude values. Since amplitudes are always non-negative, distribution $p(a_i)=\frac{1}{Z}e^{-S(a_i)}$ is assumed to be exponential i.e. $S(a_i) = \lambda a_i$. Due to this assumption, the model attempts to form a data representation keeping complex amplitudes maximally independent across subspaces, while still allowing dependence between coordinates $s^{\mathfrak{R}}, s^{\mathfrak{I}}$ which determine position within each subspace. The posterior over coefficients $s$ becomes
\begin{equation}
\label{coeffPosterior}
p(s|x,A) \propto p(x|A,s)p(s)
\end{equation}

This model was introduced in \cite{CadieuOlshausen} as a model of natural image patches. Assuming $n=T$ and $\sigma = 0$, it is equivalent to 2-dimensional Independent Subspace Analysis(ISA) \cite{HyvarinenISA}.

Figure \ref{fig1} A) depicts four exemples of complex basis functions learned by from natural, speech sounds. In addition to temporal plots in Cartesian (first row) and polar (second row) coordinates each basis function is also depicted in the frequency domain (third row). Real ($A_i^{\mathfrak{R}}$ - black lines) and imaginary ($A_i^{\mathfrak{I}}$ - gray lines) parts of basis functions do not resemble each other and are not temporally localized, capturing the non-local strucutre of speech sounds.

\subsection{Phase and amplitude continuity priors}

The sparse coding model described in the previous subsection, does not constrain the basis functions in any way. They are allowed to vary freely during the learning process. As visible on figure \ref{fig1} A), an unconstrained adaptation to natural sound corpus yields complex basis functions invariant to numerous stimulus aspects such as frequency or time shifts, not necessarily phase.

Learning a structured dictionary requires therefore placing priors over basis functions, which favour solutions of desired properties such as phase-invariance. Real and imaginary parts of a phase-shift invariant basis function, have equal, unimodal frequency spectra and both span the same temporal interval. Additionally, the imaginary part should be shifted in time a quarter of the cycle with respect to the real one.

Before proposing a prior promoting such solutions, one should note that each basis function $A_i(t)$ can be represented in polar coordinates as
\begin{equation}
\label{bfPolar}
A_i(t)=a^A_i(t)\Big(\cos \phi^A_i(t) + j\sin \phi^A_i(t)\Big) 
\end{equation}
where $a_i^A(t)$ and $\phi_i^A(t)$ denote instantaneous phase and amplitude, respectively. Angular frequency can be defined as a temporal derivative of instantaneous phase. If phase dynamics are highly variable and non-monotonic over time, real and imaginary components of this signal have non-identical spectra and/or their frequencies change in time (see figure \ref{fig1} A), second and third rows). On the other hand, by enforcing phase $\phi_i^A(t)$ to change smoothly and monotonically, one should obtain real and imaginary parts with matching frequency spectra. In the limiting case, when phase is a linear function of time, real and imaginary parts oscillate in the same frequency and are in a quadrature phase relationship.
Furthermore, vectors which span a phase-shift invariant subspace should have the same temporal support, implying that the complex amplitude should also vary slowly in time. 

In order to learn a phase-shift invariant representation of natural sounds, the present work proposes a prior over basis functions of the following form:
\begin{equation}
\label{bfPrior}
p(A_i)=p_{\phi}(A_i) p_{a}(A_i)=\frac{1}{Z} e^{-(\gamma S_{\phi}(A_i) + \beta S_a(A_i))}
\end{equation} 
Function $S_a(A_i)$ introduces the penalty proportional to the variance of amplitude's temporal derivative:
\begin{equation}
\label{samp}
S_a(A_i)=\sum_{t>1}^T \Big( \Delta a^A_i(t)\Big) ^2
\end{equation} where  $\Delta a^A_i(t) = a^A_i(t)-a^A_i(t-1)$. It promotes basis functions with a slowly-varying envelope, highly correlated between consecutive time steps. 
Phase prior $S_{\phi}$ is defined by function $S_{\phi}(A_i)$ of the following form:
\begin{equation}
\label{sphi}
S_{\phi}(A_i)= -\sum_{t>1}^T sgn\Big(\Delta \phi_i(t)\Big) \Big( \Delta \phi_i(t) \Big)^2
\end{equation}
where $\Delta \phi^A_i(t) = \phi^A_i(t)-\phi^A_i(t-1)$ and $sgn$ denotes the sign function. Similarly to $S_a(A_i)$ it promotes temporal slowness of phase. The additional factor $-sgn(\Delta \phi_i(t))$ enforces $\phi(t)$ to be larger than $\phi(t-1)$. In this way, it prevents phase from changing direction and causes it to be a non-increasing function of time. One could also enforce this by bounding the phase derivative from above: $\Delta \phi_i(t) < \Theta$. This method would however require the hand tuning of the $\Theta$ parameter. The posterior over basis functions given a data sample $x$ and its representation $s$ becomes:
\begin{equation}
\label{posteriorBf}
p(A|x,s) \propto p(x|A,s)p(A)
\end{equation}

where the likelihood model $p(x|A,s)$ is defined by equation \eqref{likelihood}. Taken together, prior $p(A_i)$ biases the learning process towards temporally localized basis functions with real and imaginary parts of the same instantaneous frequency.

Exemplary complex features learned with introduced priors are depicted on figure \ref{fig1} B). Compared with unconstrained subspaces from figure \ref{fig1} A), their amplitudes are smooth, and their phases change monotonically. Moreover, frequency spectra of $A_i^{\mathfrak{R}}$ and $A_i^{\mathfrak{I}}$ align well.
Such bases form a phase invariant representation of the data.

\subsection{Learning and inference}

Inference i.e. estimation of coefficients $s$ given a data sample $x$ is performed by finding a maximum a posteriori estimate (MAP). This corresponds to finding a mode of the posterior distribution \eqref{coeffPosterior} and can be computed via a gradient descent on the negative log-posterior (for gradient derivations please refer to the supplementary material):
\begin{equation}
\label{Es}
E_s = \frac{1}{2\sigma^2} \left[\sum_{t=1}^T\Big(\hat{x}(t)-x(t)\Big)^2\right] + \lambda  \sum_{i=1}^{n/2} S(a_i)
\end{equation}

Learning of basis functions $A$ is performed in two steps as in\cite{OlshausenField, CadieuOlshausen}. Firstly, coefficients $s$ are inferred given the data sample $x$. Secondly, using inferred $s$ values, a gradient update is performed on basis functions shifting solutions towards the mode of the posterior distribution \eqref{posteriorBf}. This is equivalent to minimization of the following energy function: 
\begin{equation}
\label{Ebf}
E_A = \frac{1}{2\sigma^2} \left[\sum_{t=1}^T\Big(\hat{x}(t)-x(t)\Big)^2\right] + \gamma \sum_{i=1}^{n/2}S_{\phi}(A_i)+ \beta \sum_{i=1}^{n/2}S_{a}(A_i)
\end{equation} where $\gamma \in \mathbb{R}$ and $\beta \in \mathbb{R}$ are free parameters which control the strength of each prior. They are set to be smaller than one to prevent from dominating over the error term. This prevents the model from learning trivial or non-data matching solutions (e.g. very low frequencies). The two steps are iterated until the algorithm converges. After every learning iteration, real and imaginary vectors spanning each subspace are orthogonalized by Gram-Schmidt orthogonalization.

The error term in the equation \eqref{Ebf} as well as amplitude and phase penalty terms $S_{a}$ and $S_{\phi}$ usually have very different numerical values. For this reason, at every iteration of the learning process each term contributing to the gradient of function $E_A$ was normalized to the unit length. Prior-related terms are then multiplied by smaller than $1$, strength-controlling constants $\gamma \in [0, 1]$ and $\beta \in [0, 1]$. In this way, gradient information was used to find an appropriate direction in the search space while preventing terms with larger numerical values from dominating the search.

\section{Properties of representations learned using speech sounds}
\label{section3}

Natural sound statistics were analysed by learning complex dictionaries and analysing properties of obtained basis functions. Complete and twice overcomplete representations were learned with and without basis function priors. This resulted in the total number of four dictionaries. Due to space constraints, entire dictionaries are visualized in time and frequency domain in the supplementary material.

All models were trained using a speech corpus from the International Phonetic Database Handbook \cite{IPA}. This dataset contains sounds of human speakers telling a story in $27$ different languages. Speech comprises a variety of acoustic structures, both harmonic and non-harmonic. One should note that it may not perfectly reflect environmental sound statistics, however it has been used before as a proxy for natural sounds \cite{Carlson, Terashima, SmithLewicki}.
All sound files were down sampled to $8000$ Hz from their original sampling rates. For training, $50000$ intervals were randomly sampled from all recording files. Each interval was $128$ samples long which corresponds to $16$ ms. Prior to training, $18$ principal components of the data, explaining $0.001$ of total variance, were rejected. This corresponds to low-pass filtering the data with the $3500$ Hz cutoff frequency. The sparsity control parameter $\lambda$ was set to $0.1$ value. After multiple experiments, the prior strengths $\gamma$ and $\beta$ were set to $0.2$ and $0.1$, respectively. 

\begin{figure}[ht]
\label{fig2}
\begin{center}
\includegraphics{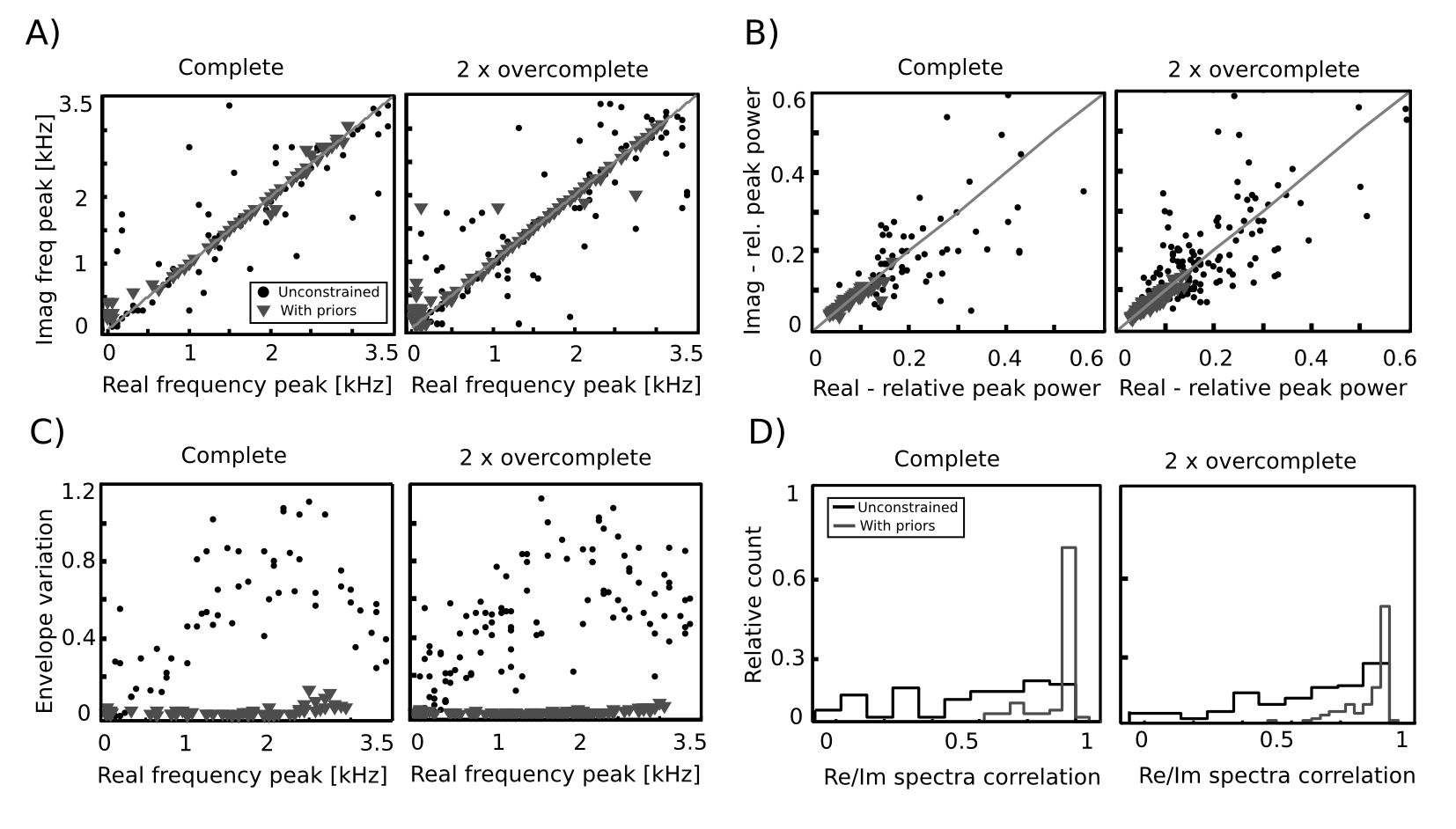}
\end{center}
\caption{A) Spectral peaks of real and imaginary vectors B) Concentration coefficients of real and imaginary vectors C) Envelope variation plotted against peak frequency of the real part D) Distributions of spectra correlations}
\end{figure}

Spectra of real and imaginary parts were compared by plotting their peaks against each other, and computing correlations. Spectral peaks of unconstrained basis functions do not match well (figure \ref{fig2} A), black circles). They are broadly scattered around the diagonal in both - complete and overcomplete case. Each such complex basis function, consists of two vectors of different frequency. This is in contrast to basis functions learned with priors (gray triangles), which are concentrated around diagonals, with only a few exceptions. This means that spectra of their real and imaginary parts have the same frequency peak.

To go beyond peak comparison, and to test how well basis functions aligned in the frequency domain, correlations between their spectra were computed. Correlation equal to $1$ means that real and imaginary parts covaried together, and strongly overlapped, while a low correlation value implies highly different spectra. Normalized correlation values were histogramed and are depicted on figure \ref{fig2} D). A clear difference between prior based (gray lines) and unconstrained (black lines) dictionaries is visible. Correlation distributions of the latter ones are quite broad (although in the overcomplete case a stronger peak close to $1$ is present) and include all possible values. Spectrum correlations of prior based basis functions, are in turn, strongly concentrated around the maximal value i.e. $1$ implying similarity of real and imaginary parts.

Spectral breadth of basis functions was assessed by computing the concentration index (CI) i.e. ratio between the peak value of the spectrum and the total spectral power. CI quantifies how well are the basis functions localized in the frequency domain. This measure was used instead of computing bandwidth, since spectra of some basis functions consisted of two or more isolated peaks and bandwidth is not well defined in such cases. CIs of real and imaginary vectors are plotted against each other on figure \ref{fig2} B). Unconstrained basis functions (black circles) tend to cluster along the diagonal for low CI values and for higher ones they diverge. This means, that if either - real or imaginary vector has a pronounced, strongly localized peak in the frequency spectrum other one will be more broad. Prior-based basis functions lie along the diagonal, meaning that spectra of their components have similar degree of concentration. They are, however, much more broad than unconstrained basis functions, with CIs not exceeding the $0.2$ value.

\begin{figure}[ht]
\label{fig3harm}
\begin{center}
\includegraphics[width=\textwidth]{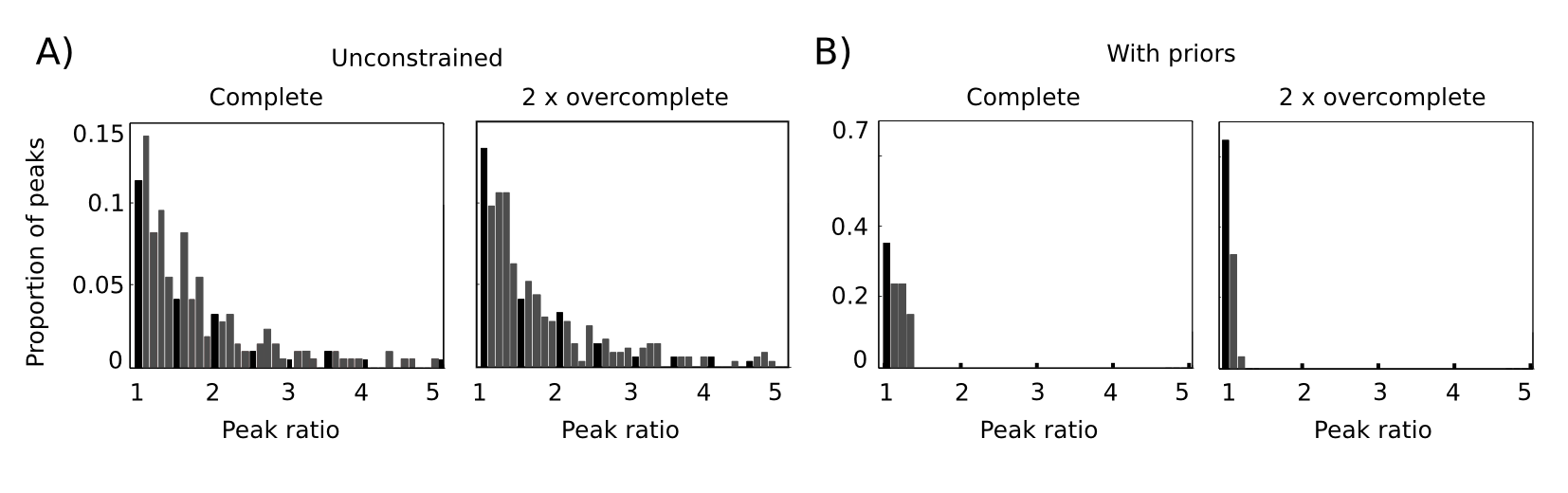}
\end{center}
\caption{Peak ratio distributions. A) unconstrained dictionaries. B) prior based dictionaries. Multiplications of $0.5$ are marked black.}
\end{figure}

It has been suggested that statistical models of natural sounds capture harmonic frequency relationships \cite{Terashima, Terashima2}. The harmonic structure of natural sounds gives rise to non-local correlations in the frequency domain, which is different from local pixel correlations of natural image patches. To test whether cross-frequency couplings learned by the unconstrained model reflects harmonic strucutre of speech the following analysis was performed. Firstly each real and imaginary pair of vectors was converted into frequency domain. Then, for each pair spectral peaks were extracted (a frequency was defined as a peak if it contained more than $0.1$ of the total spectral power). Ratios of the minimal peak value to remaining values were computed. Ratio histograms are depicted of figure \ref{fig3harm}

Prior based features contained mostly a single peak equal for real and imaginary vectors which is reflected by concentration of nearly all peak ratios at $1$. Unconstrained features revealed different structure. Sharp histogram peaks are visible directly at or very close to
multiplcations of $0.5$ (one should compare this figure to figure 4 B in \cite{Terashima2}). This is an indication that indeed, the unconstrained model has learned harmonic frequency relationships. This can be expected, since real and imaginary parts of complex valued features capture mutually dependent data aspects. 

To evaluate properties of basis functions in the temporal domain, variability of normalized amplitudes was computed, according to the equation \eqref{samp}. Results are plotted against peak frequency of the real part on figure \ref{fig2} C). In both - complete and overcomplete case, the unconstrained basis functions (black circles) reveal a frequency dependence. For higher frequencies, amplitudes vary more strongly, although around $3$ kHz variability decreases again. Amplitudes of features learned in the presence of priors is much lower, as expected. The slow amplitude prior quenches the variability almost to $0$, with a slight raise observable in higher frequency regimes.

\begin{figure}[ht]
\label{fig3}
\begin{center}
\includegraphics{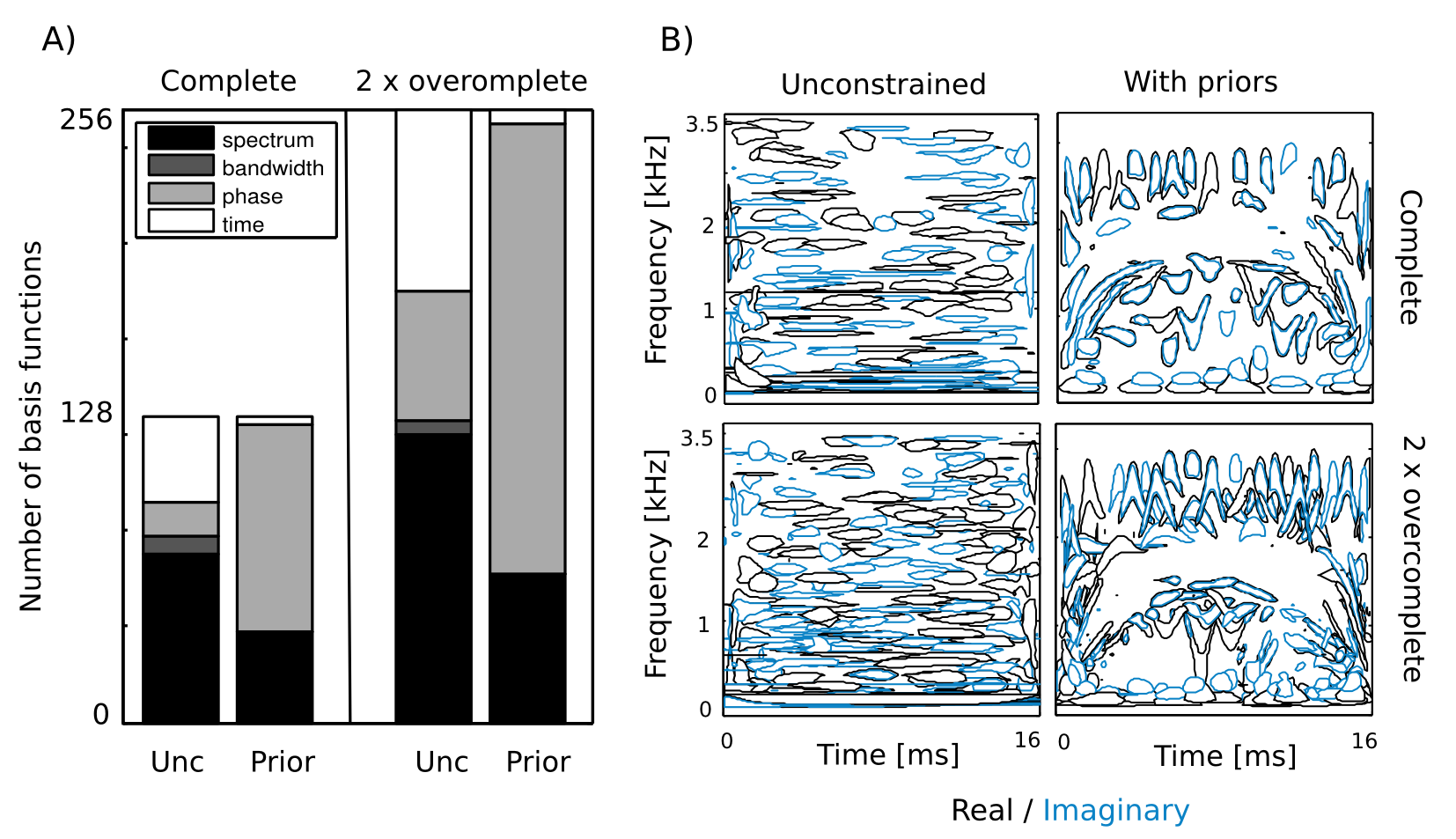}
\end{center}
\caption{A) Contour plots of basis function Wigner distributions. Black contours are real and blue are imaginary B)Proportion of invariances learned by each dictionary. Black - spectrum invariance, dark gray - bandwidth invariance, light gray - phase invariance, white - time invariance.}
\end{figure}

In order to understand how learned basis functions tile the time-frequency plane, Wigner distributions of each vector were computed. Wigner distributions describe spectrotemporal energy distribution of temporal signal. In the next step, equiprobability contours corresponding to $0.7$ probability value for real and $0.8$ for imaginary parts were plotted (figure \ref{fig3}, black and blue contours respectively). Such representation of temporal basis functions on the time frequency plane was introduced by \cite{Abdallah}. If both vectors spanning each subspace were indeed phase invariant, their equiprobability contours should lie within each other. Figure \ref{fig3} shows, that this is rarely the case for unconstrained basis functions (first column). While they tile the time-frequency plane uniformly, corresponding real and imaginary parts often lie far apart, modelling different regions.
Prior-constrained basis functions can not vary their real and imaginary parts independently during the learning process. While unconstrained bases increase their temporal support with decreasing frequency, ones learned with priors are strongly localized in time, independent of their spectrum. In most cases, corresponding real and imaginary vectors occupy the same area, however a tendency is visible among imaginary components of low-frequency features to be elevated with respect to their real counterparts along the frequency axis.
An interesting effect is visible in the region between $0.5$ and $2$ kHz. There, constrained basis functions become broadband and span large frequency intervals. In some cases the bandwidth reversal occurs. This is visible as ''banana-like'' shapes. The empty regions above and below, are covered by equiprobability contours corresponding to probabilities lower than $0.7$, which are not visible on the plot.
These structures reflect temporal frequency variation of the basis functions (see figure 1 B, second row - phases are monotonic, but rather piecewise linear functions of time). A possible explanation of their emergence is that real and imaginary vectors tend to diverge from each other (as in the unconstrained model), but the prior forces them to stay close on the time frequency plane. Interestingly a qualitative change in Wigner distribution shapes was observed in this frequency regime by Abdallah and Plumbey \cite{Abdallah}, who studied independent components of natural sounds. Such behaviour may imply, that data drives real and imaginary parts of basis functions to span distant frequencies, while priors ''keep them together''.
Figure \ref{fig3} A confirms that constrained and unconstrained representations differ strongly in their spectro-temporal structure.

As mentioned before each complex basis function forms an invariant representation of some data feature i.e. varying phase within subspace spanned by vectors $A^{\mathfrak{R}}$ and $A^{\mathfrak{I}}$ generates features of different quality, while keeping the complex amplitude constant. To summarize stimulus-invariances captured by every dictionary, basis functions were assigned to four classes of invariance, according to the following criteria:
\begin{enumerate}
\item Spectral invariance - frequency peaks of real and imaginary vectors differ by more than $100$~Hz
\item Bandwidth invariance - real and imaginary parts have the same frequency peak, but different concentration index CI
\item Time invariance - frequency peaks and concentrations match, real and imaginary vectors are shifted in time more than a period of peak frequency
\item Phase invariance - same as above, with a shift smaller than the peak frequency period.
\end{enumerate}

Figure \ref{fig3} B) depicts how many basis functions from each dictionary fall within each invariance class. Unconstrained representations capture mostly spectral invariances (black color). Temporal invariances (white) are the second largest class, while phase invariant features (light gray) constitute a minor fraction $11 \%$ in complete and $21 \%$ in the overcomplete case.
In contrast, representations learned with phase and amplitude priors capture mostly phase invariances - $67 \%$ and $73 \%$ respectively. One should note, that spectral invariances learned by constrained models are much less diverse than ones captured by unconstrained ones. They result mostly from the slight misalignment of spectral peaks - see figure \ref{fig2} A).

\begin{figure}[ht]
\label{fig5}
\begin{center}
\includegraphics{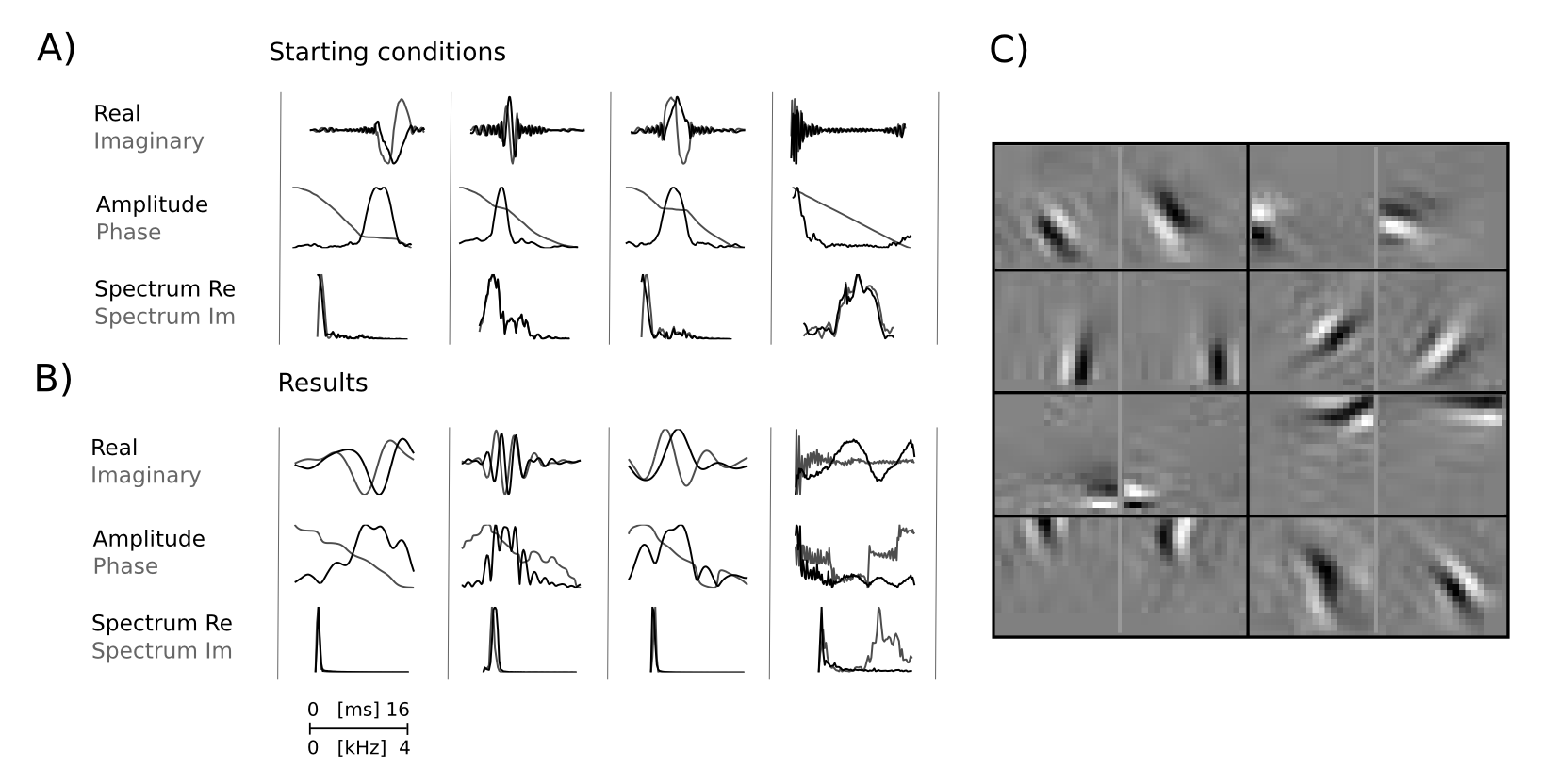}
\end{center}
\caption{Control experiments. A) Initial conditions - basis functions learned with smoothness priors. B) The same basis functions after $30000$ learning iterations. C) Complex basis functions learned without priors from the image data.}
\end{figure}

In order to reassure that the structure of the unconstrained basis functions does not constitute a learning artifact, two control experiments were performed. In the first one, a complete dictionary of prior based basis functions was used as an initialization for unconstrained learning. If features revealing complex invariances were a robust data property not merely a local minimum, the resulting dictionary should differ from the original one. This is indeed what happened. Figure \ref{fig5} A depicts four randomly selected, prior based basis functions used as initial conditions. Figure \ref{fig5} B shows the same basis functions after $30000$ learning iterations. Structure induced by priors has vanished. Spectra of single vectors are more localized, but not necessarily match each other (see the last basis function). Phases and amplitudes vary strongly in time. This observation confirms that structure of the unconstrained dictionary forms a representation better spanning the data space. Secondly, unconstrained features were learned from natural image dataset (taken from \cite{HyvBook}). Eight randomly selected subspaces are depicted on figure \ref{fig5} C. As expected, they are localized in space and frequency and are in a quadrature phase relationship.

\section{Performance of learned representations}
\label{section4}

In order to compare how well different dictionaries model the underlying data distribution, two different criteria were used. Firstly, performance of a dictionary in a denoising task was measured and secondly relative coding efficiency was compared by estimating the entropy of linear coefficients. Both tests were performed using a testing dataset of $20000$ samples drawn from the IPA speech corpus. The test dataset was preprocessed in the same way as the training one.

\subsection{Denoising}

The ability of different dictionaries to match a typical structure in the data in the presensce of noise was quantified. This may be also understood as an indirect estimate of the likelihood, since it is known that models of high-likelihood perform well in denoising tasks \cite{Zoran}.

Each vector $x$ from the test dataset was blurred with an i.i.d. gaussian noise with sd $\sigma = 0.1$. Coefficients $s$ were inferred from the noisy sample. In the next step Peak Signal to Noise Ratio (PSNR) defined as $PSNR = 20 \log_{10}(\frac{1}{\|x-\hat{x}\|})$ (where $\hat{x}$ is the reconstruction of the data vector given inferred $s$) was computed.

\begin{figure}[ht]
\label{fig4}
\begin{center}
\includegraphics{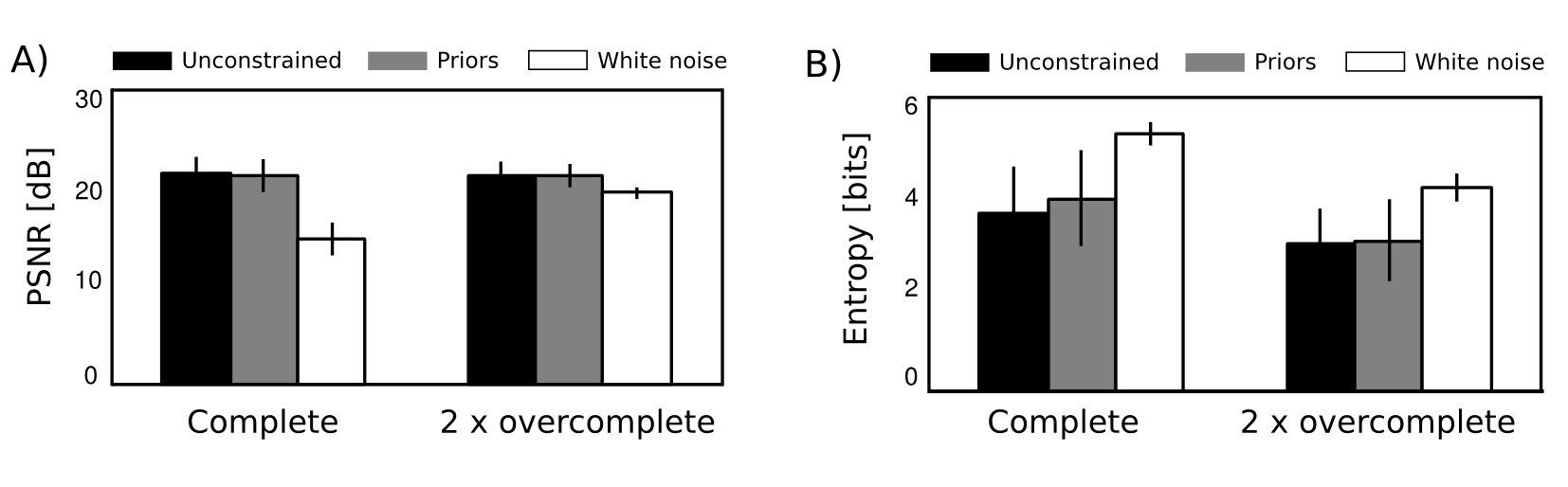}
\end{center}
\caption{A) Average performance in denoising task measured by PSNR in dB. Vertical bars indicate standard deviation. B) Average coefficient entropy estimated from histograms.}
\end{figure}

Average PSNR for each dictionary is plotted on figure \ref{fig4} A). For comparison, denoising was also performed using a basis consisting of orthogonalized white noise vectors. Surprisingly, all dictionaries adapted to the data showed very similar performance. Priors presence and overcompletness did not affect results much. Despite different basis function shapes dictionaries were able to reconstruct orginal sound chunks, while rejecting the noise. White noise bases gave, as expected, lower reconstruction quality, which has however risen in the overcomplete case. Such behavior is expected, since the model had more degrees of freedom to match the data structure.

\subsection{Quantifying coding efficiency}

One way to asess coding efficiency given a fixed dictionary, is to estimate entropies of linear coefficients $s^{\mathfrak{R}}$ and $s^{\mathfrak{I}}$ \cite{LewickiOlshausen}. By Shannon's source coding theorem, the entropy of the data distribution $p$ constitutes a lower bound on code-word length: 
\begin{equation}
\label{entropy}
\mathcal{L} \geq H(p) = -\sum_x p(x) \log p(x)
\end{equation}
If the true probability distribution $p$ is unknown and is approximated by a proxy distribution $q$, the average code length becomes:
\begin{equation}
\label{entropy2}
\mathcal{L} \geq H(p) + D_{KL}(p\|q)
\end{equation}
where $D_{KL}(p\|q)$ is the Kullblack-Leibler divergence from distribution $p$ to $q$. Therefore, representation which yields lower code length should be closer to the true underlying distribution of the data. For details see \cite{LewickiSejnowski, LewickiOlshausen}.

Coefficient entropies were estimated by creating normalized histograms with $256$ bins. Entropy was estimated as $\hat{H}(p)=-\sum_i \frac{n_i}{N} \log_2 \frac{n_i}{N}$, where $n_i$ is a number of counts in $i$-th bin and $N$ is the  number of samples. One should note, that even though the estimation was strongly biased, the goal was to use entropy as a relative measure of coding efficiency using different representations, not as an absolute one.

Average coefficient entropy for all learned dictionaries is plotted on figure \ref{fig4} B). Overall, overcomplete representations yielded lower entropies. It means that models with more sparse causes than dimensions explain the data better and may be surprising when compared with natural image models, where overcomplete representations yield higher coefficient entropies \cite{LewickiOlshausen}. Representing the data using phase-invariant basis functions required slightly more bits than using unconstrained ones. This indicates that even though both representations gave good denoising performance, the unconstrained model may be closer to the true, data generating distribution. As expected, white noise bases required more bits per coefficient to encode speech sounds and their overall performance was poor.

\section{Summary}
This work has addressed two lines of research. Firstly natural sound statistics were studied by learning sparse complex valued representations and analysis of obtained features. It has been demonstrated that in contrary to natural images learned features are invariant to many different stimulus aspects, not only phase. One should note that phase in short sound waveforms is a very different physical quantity than spatial phase of natural image patches. It can be expected that temporal structure of air pressure waveforms is going to have different statistical properties than spatial structure of reflected light. The present work shows that intuitions (phase invariance of sparse, complex dictionaries) gained from learning representations in one signal domain (images) may not transfer to others (e.g. sound). Present findings go along recent results by Terashima and colleagues \cite{Terashima, Terashima2}, who suggested that differences in spatial organization of visual and auditory cortices may reflect different correlation structures of natural sounds and images.

In parallel to analysis of sound statistics, priors promoting phase invariant sparse codes were proposed. The form of prior which penalizes temporal variability of the signal is long known \cite{Foldiak, Wiskott, HyvarinenBubbles, TurnerSFA}. Recently, it was used to learn complex, temporally persistent representations from sequences of natural image patches \cite{CadieuOlshausenInv}. Here slow priors were placed on amplitudes and phases of basis functions, not coefficients. Phase prior includes also an additional term promoting monotonic change of phase in time. 

Complete and overcomplete speech representations were learned using unconstrained and prior-based models. Obtained basis functions highly differ in shape and spectro-temporal properties. Despite differences, they perform equally well in a denoising task, and yield similar coefficient entropies. This implies that prior based dictionaries can be used without quality loss to represent natural sounds in tasks such spatial hearing, where phase information has to be made explicit.

Proposed approach to learn complex dictionaries may find applications outside natural scene statistics research. For instance, phase invariant dictionaries are useful in modelling time-epoched signal, where epochs can be misaligned \cite{Hitziger}.

Many other prior forms may be selected to learn structured signal representations. For instance penalizing variability of the second temporal phase derivative should yield basis functions which are well localized in frequency. Applicability and usefulness of such prior in learning efficient representations of sensory data is a subject for future work.

\subsubsection*{Acknowledgments}
This work was funded by the DFG Graduate College ``Interneuro``

\bibliographystyle{abbrv}
{\footnotesize 
\bibliography{iclr}

\begin{thebibliography}{10}

\bibitem{Abdallah}
S.~A. Abdallah and M.~D. Plumbley.
\newblock If the independent components of natural images are edges, what are
  the independent components of natural sounds.
\newblock In {\em Proceedings of International Conference on Independent
  Component Analysis and Signal Separation (ICA2001)}, pages 534--539, 2001.

\bibitem{IPA}
I.~P. Association.
\newblock Handbook of the international phonetic association: A guide to the
  use of the international phonetic alphabet.

\bibitem{BellSejnowski}
A.~J. Bell and T.~J. Sejnowski.
\newblock Learning the higher-order structure of a natural sound*.
\newblock {\em Network: Computation in Neural Systems}, 7(2):261--266, 1996.

\bibitem{CadieuOlshausenInv}
C.~F. Cadieu and B.~A. Olshausen.
\newblock Learning intermediate-level representations of form and motion from
  natural movies.
\newblock {\em Neural computation}, 24(4):827--866, 2012.

\bibitem{Carlson}
N.~L. Carlson, V.~L. Ming, and M.~R. DeWeese.
\newblock Sparse codes for speech predict spectrotemporal receptive fields in
  the inferior colliculus.
\newblock {\em PLoS computational biology}, 8(7):e1002594, 2012.

\bibitem{Foldiak}
P.~F{\"o}ldi{\'a}k.
\newblock Learning invariance from transformation sequences.
\newblock {\em Neural Computation}, 3(2):194--200, 1991.

\bibitem{Grosse}
R.~Grosse, R.~Raina, H.~Kwong, and A.~Y. Ng.
\newblock Shift-invariance sparse coding for audio classification.
\newblock {\em arXiv preprint arXiv:1206.5241}, 2012.

\bibitem{Hitziger}
S.~Hitziger, M.~Clerc, A.~Gramfort, S.~Saillet, C.~B{\'e}nar, and
  T.~Papadopoulo.
\newblock Jitter-adaptive dictionary learning-application to multi-trial
  neuroelectric signals.
\newblock {\em International Conference on Learning Representations}, 2013.

\bibitem{Hsu}
A.~Hsu, S.~M. Woolley, T.~E. Fremouw, and F.~E. Theunissen.
\newblock Modulation power and phase spectrum of natural sounds enhance neural
  encoding performed by single auditory neurons.
\newblock {\em The Journal of neuroscience}, 24(41):9201--9211, 2004.

\bibitem{HyvarinenISA}
A.~Hyv{\"a}rinen and P.~Hoyer.
\newblock Emergence of phase-and shift-invariant features by decomposition of
  natural images into independent feature subspaces.
\newblock {\em Neural Computation}, 12(7):1705--1720, 2000.

\bibitem{HyvBook}
A.~Hyv{\"a}rinen, J.~Hurri, and P.~O. Hoyer.
\newblock {\em Natural Image Statistics}, volume~39.
\newblock Springer, 2009.

\bibitem{HyvarinenBubbles}
A.~Hyv{\"a}rinen, J.~Hurri, and J.~V{\"a}yrynen.
\newblock Bubbles: a unifying framework for low-level statistical properties of
  natural image sequences.
\newblock {\em JOSA A}, 20(7):1237--1252, 2003.

\bibitem{Laparra}
V.~Laparra, M.~U. Gutmann, J.~Malo, and A.~Hyv{\"a}rinen.
\newblock Complex-valued independent component analysis of natural images.
\newblock In {\em Artificial Neural Networks and Machine Learning--ICANN 2011},
  pages 213--220. Springer, 2011.

\bibitem{Lewicki}
M.~S. Lewicki.
\newblock Efficient coding of natural sounds.
\newblock {\em Nature neuroscience}, 5(4):356--363, 2002.

\bibitem{LewickiOlshausen}
M.~S. Lewicki and B.~A. Olshausen.
\newblock Probabilistic framework for the adaptation and comparison of image
  codes.
\newblock {\em JOSA A}, 16(7):1587--1601, 1999.

\bibitem{LewickiSejnowski}
M.~S. Lewicki and T.~J. Sejnowski.
\newblock Learning overcomplete representations.
\newblock {\em Neural computation}, 12(2):337--365, 2000.

\bibitem{CadieuOlshausen}
B.~A. Olshausen, C.~F. Cadieu, and D.~K. Warland.
\newblock Learning real and complex overcomplete representations from the
  statistics of natural images.
\newblock In {\em SPIE Optical Engineering+ Applications}, pages
  74460S--74460S. International Society for Optics and Photonics, 2009.

\bibitem{OlshausenField}
B.~A. Olshausen and D.~J. Field.
\newblock Sparse coding with an overcomplete basis set: A strategy employed by
  v1?
\newblock {\em Vision research}, 37(23):3311--3325, 1997.

\bibitem{SmithLewickiNC}
E.~Smith and M.~S. Lewicki.
\newblock Efficient coding of time-relative structure using spikes.
\newblock {\em Neural Computation}, 17(1):19--45, 2005.

\bibitem{SmithLewicki}
E.~C. Smith and M.~S. Lewicki.
\newblock Efficient auditory coding.
\newblock {\em Nature}, 439(7079):978--982, 2006.

\bibitem{SmithChimeras}
Z.~M. Smith, B.~Delgutte, and A.~J. Oxenham.
\newblock Chimaeric sounds reveal dichotomies in auditory perception.
\newblock {\em Nature}, 416(6876):87--90, 2002.

\bibitem{Terashima2}
H.~Terashima and H.~Hosoya.
\newblock Sparse codes of harmonic natural sounds and their modulatory
  interactions.
\newblock {\em Network: Computation in Neural Systems}, 20(4):253--267, 2009.

\bibitem{Terashima}
H.~Terashima and M.~Okada.
\newblock The topographic unsupervised learning of natural sounds in the
  auditory cortex.
\newblock In {\em Advances in Neural Information Processing Systems 25}, pages
  2321--2329, 2012.

\bibitem{Tosic}
I.~Tosic and P.~Frossard.
\newblock Dictionary learning.
\newblock {\em Signal Processing Magazine, IEEE}, 28(2):27--38, 2011.

\bibitem{TurnerSFA}
R.~Turner and M.~Sahani.
\newblock A maximum-likelihood interpretation for slow feature analysis.
\newblock {\em Neural computation}, 19(4):1022--1038, 2007.

\bibitem{WangOlshausenMing}
J.~Wang, , B.~Olshausen, and V.~Ming.
\newblock A sparse subspace model of higher-level sound structure.
\newblock {\em COSYNE Proceedings}, 2008.

\bibitem{Wiskott}
L.~Wiskott and T.~J. Sejnowski.
\newblock Slow feature analysis: Unsupervised learning of invariances.
\newblock {\em Neural computation}, 14(4):715--770, 2002.

\bibitem{Yaghoobi}
M.~Yaghoobi, L.~Daudet, and M.~E. Davies.
\newblock Parametric dictionary design for sparse coding.
\newblock {\em Signal Processing, IEEE Transactions on}, 57(12):4800--4810,
  2009.

\bibitem{Zoran}
D.~Zoran and Y.~Weiss.
\newblock From learning models of natural image patches to whole image
  restoration.
\newblock In {\em Computer Vision (ICCV), 2011 IEEE International Conference
  on}, pages 479--486. IEEE, 2011.

\end{thebibliography}
}
\newpage

\section*{Supplementary material}
\setcounter{figure}{0}
\setcounter{equation}{0}

\subsection{Gradient derivations}
In this section learning rules i.e. gradients over linear coefficients and basis functions are derived.

\subsubsection{Coefficients gradient}

Let $\hat{x}$ be the reconstruction of the original data vector $x$
using inferred coefficients $s$ and basis functions $A$: \begin{equation}
\hat{x}(t)=\sum_{i=1}^{n/2}\mathfrak{R}\{s^*_{i}A_{i}(t)\}\label{rec}\end{equation}

Residue $r(t)$ i.e. difference between the data vector and its reconstruction
becomes: \begin{equation}
r(t)=x(t)-\hat{x}(t)\label{rec}\end{equation}

Inference of coefficients is equivalent to minimization of the following
energy function:

\begin{equation}
E_{s}=\frac{1}{2\sigma^{2}}\left(\sum_{t=1}^{T}r(t)^{2}\right)+\lambda\sum_{i=1}^{n/2}S(a_{i})\label{Es2}\end{equation}
 In the present work use of function $S(a_{i})=a_{i}$ is equivalent
to placing an $L1$ norm penalty on amplitudes $a_{i}=\|s_{i}\|=\sqrt{s_{i}^{\mathfrak{R}^{2}}+s_{i}^{\mathfrak{I}^{2}}}$.
The gradient over linear coefficients $s^{\mathfrak{R}},s^{\mathfrak{I}}$
becomes: \begin{equation}
\frac{\partial E_{s}}{\partial s_{i}^{\mathfrak{S}}}\propto\frac{1}{\sigma^{2}}\sum_{t=1}^{T}A_{i}^{\mathfrak{S}}(t)r(t)+\lambda\frac{s_{i}^{\mathfrak{S}}}{\sqrt{s_{i}^{\mathfrak{R}^{2}}+s_{i}^{\mathfrak{I}^{2}}}}\label{gradReal}\end{equation}

Where $\mathfrak{S}\in\{\mathfrak{R},\mathfrak{I}\}$ indicates whether
the coefficient is real or imaginary.

\subsubsection{Basis function gradient}

Basis functions are learned by performing a gradient step given inferred
$s$ values. The negative log-posterior is given by: \begin{equation}
E_{A}=E_{Res}+\gamma E_{\phi}+\beta E_{Sa}=\frac{1}{2\sigma^{2}} \left( \sum_{t=1}^{T}r(t)^{2} \right)+\gamma\sum_{i=1}^{n/2}S_{\phi}(A_{i})+\beta\sum_{i=1}^{n/2}S_{a}(A_{i})\label{Ebf2}\end{equation}

Functions $S_{\phi}(A_{i})$ and $S_{a}(A_{i})$ are of following
forms:

\begin{equation}
S_{a}(A_{i})=\sum_{t>1}^{T}\Big( \Delta a_{i}^{A}(t)\Big) ^{2}\label{samp2}
\end{equation}

\begin{equation}
S_{\phi}(A_{i})=-\sum_{t>1}^{T}sgn \Big( \Delta\phi_{i}(t) \Big) \Big( \Delta\phi_{i}(t) \Big) ^{2}\label{sphi2}
\end{equation}

where $\Delta a_{i}^{A}(t)=a_{i}^{A}(t)-a_{i}^{A}(t-1)$ and $\Delta\phi_{i}^{A}(t)=\phi_{i}^{A}(t)-\phi_{i}^{A}(t-1)$.
Priors defined by $S_{a}$ and $S_{\phi}$ temporal phase and amplitude
correlations respectively.

Gradient of equation \ref{Ebf2} can be decomposed into three terms:
\begin{equation}
\frac{\partial}{\partial A_{i}(t)}E_{A}=\frac{\partial}{\partial A_{i}(t)}E_{Res}+\beta\frac{\partial}{\partial A_{i}(t)}E_{Sa}+\gamma\frac{\partial}{\partial A_{i}(t)}E_{S\phi}\label{bfGradDecomp}\end{equation}
 representing the reconstruction error term and phase and amplitude
priors consecutively. In polar coordinates, for $1<t<T$ phase prior
gradient is: 
\begin{equation}
\label{term3}
\frac{\partial}{\partial\phi_{i}^{A}(t)}\propto2\phi_{i}^{A}(t)\Big[ sgn \Big( \Delta\phi_{i}^{A}(t+1) \Big) \phi_{i}^{A}(t+1)-sgn \Big( \Delta\phi_{i}^{A}(t) \Big) \phi_{i}^{A}(t) \Big] 
\end{equation}
 For boundary conditions i.e. $t=1$ and $t=T$, this gradient becomes
consecutively: 
\begin{equation}
\label{term3t1}
\frac{\partial E_{\phi}}{\partial\phi_{i}^{A}(1)}E_{\phi}\propto2\phi_{i}^{A}(1)sgn \Big( \Delta\phi_{i}^{A}(2) \Big) \phi_{i}^{A}(2)
\end{equation}

\begin{equation}
\label{term3tT}
\frac{\partial E_{\phi}}{\partial\phi_{i}^{A}(T)}\propto-2\phi_{i}^{A}(T)sgn \Big( \Delta\phi_{i}^{A}(T) \Big) \phi_{i}^{A}(T)
\end{equation}

In the same way, the amplitude term gradient is defined separately
for $1<t<T$: 
\begin{equation}
\label{amp1}
\frac{\partial E_{a}}{\partial a_{i}^{A}(t)}\propto2 \Big( \Delta a_{i}^{A}(t)-\Delta a_{i}^{A}(t+1) \Big)
\end{equation}
 and separately for the boundary conditions ($t=1$ and $t=T$):

\begin{equation}
\label{amp2}
\frac{\partial E_{a}}{\partial a_{i}^{A}(1)}\propto-2\Delta a_{i}^{A}(2)
\end{equation}

\begin{equation}
\label{amp3}
\frac{\partial}{\partial a_{i}^{A}(T)}E_{a}\propto2\Delta a_{i}^{A}(T)
\end{equation}

The residue term is most conveniently represented in Cartesian coordinates
for real and imaginary coefficients $s_{i}^{\mathfrak{S}}$, where,
as previously, $\mathfrak{S}\in\{\mathfrak{R},\mathfrak{I}\}$, indicates
whether coefficient is real or imaginary: \begin{equation}
\label{amp4}
\frac{\partial E_{Res}}{\partial A_{i}^{\mathfrak{S}}(t)}\propto\frac{s_{i}^{\mathfrak{S}}}{\sigma^{2}}r(t)
\end{equation}

\clearpage{} 
\subsection{Dictionary plots}
\begin{figure}[ht]
\begin{center}
\includegraphics[height=0.85\textheight]{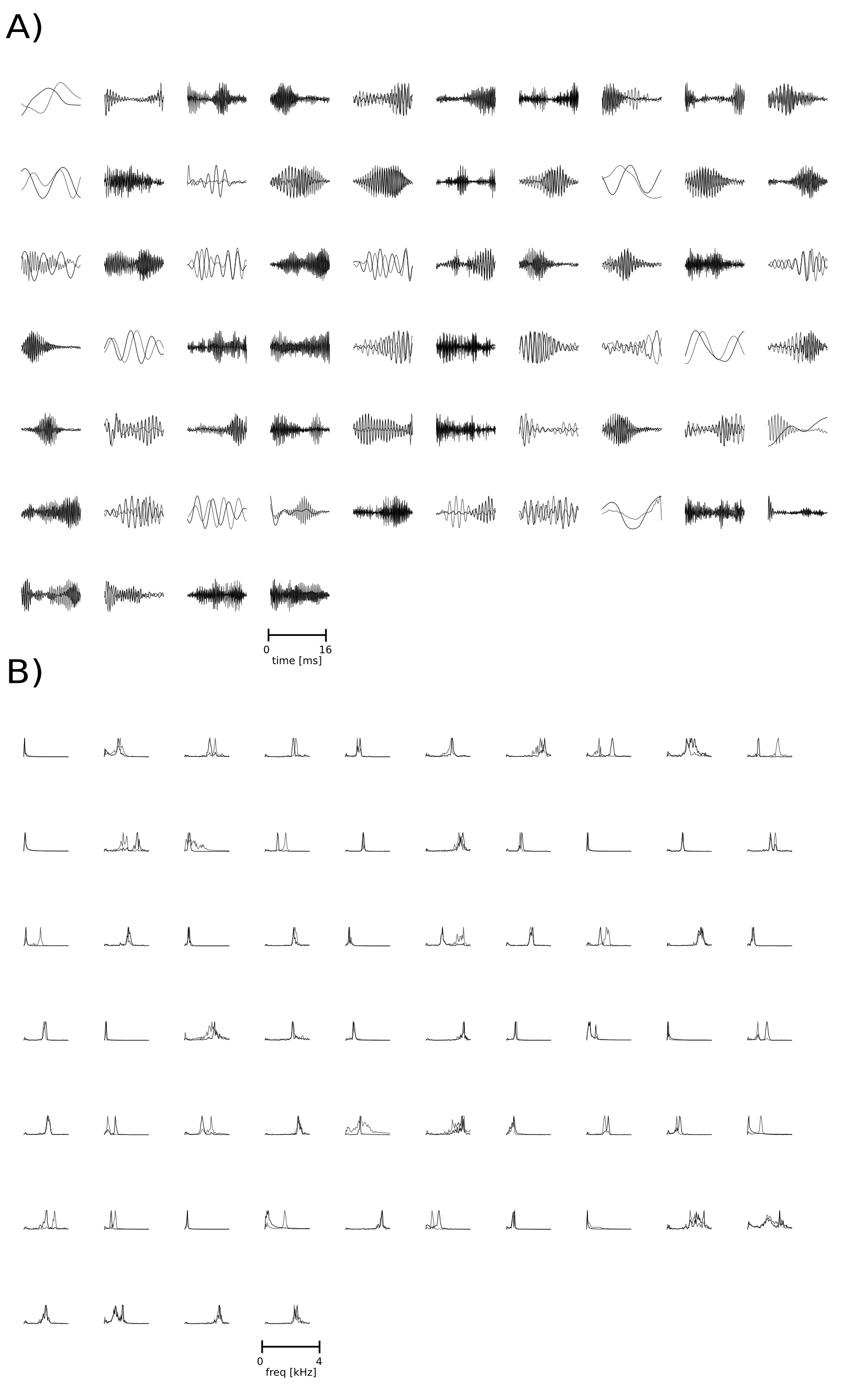}
\end{center}
\caption{Complete set of complex, unconstrained basis functions. Black lines depict real parts, and gray - imaginary ones. A) Basis functions in temporal domain B) Basis function in frequency domain}
\end{figure}

\begin{figure}[h]
\begin{center}
\includegraphics[height=0.85\textheight]{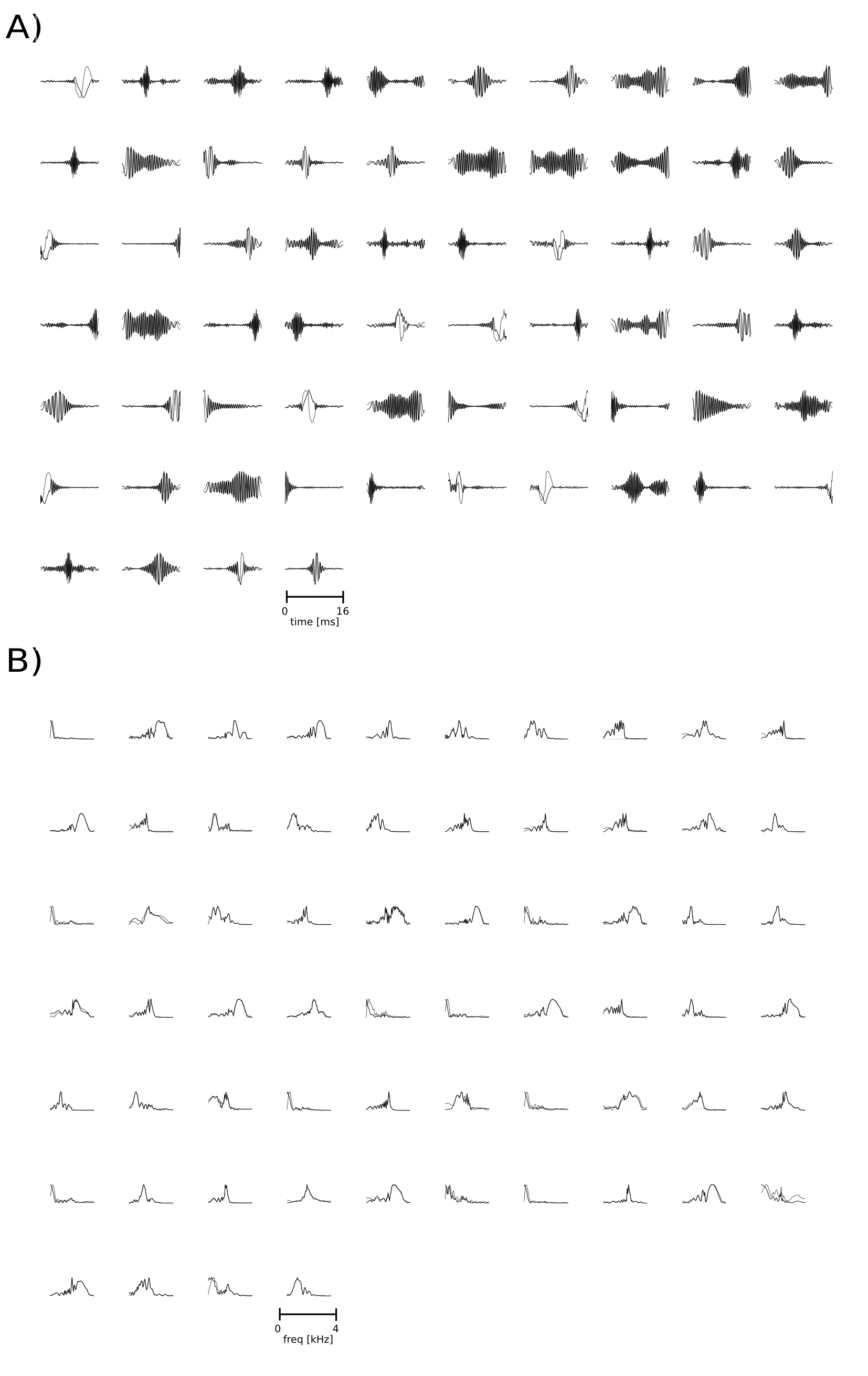}
\end{center}
\caption{Complete set of complex, basis functions learned with priors. Black lines depict real parts, and gray - imaginary ones. A) Basis functions in temporal domain B) Basis function in frequency domain}
\end{figure}

\begin{figure}[h]
\begin{center}
\includegraphics[height=0.85\textheight]{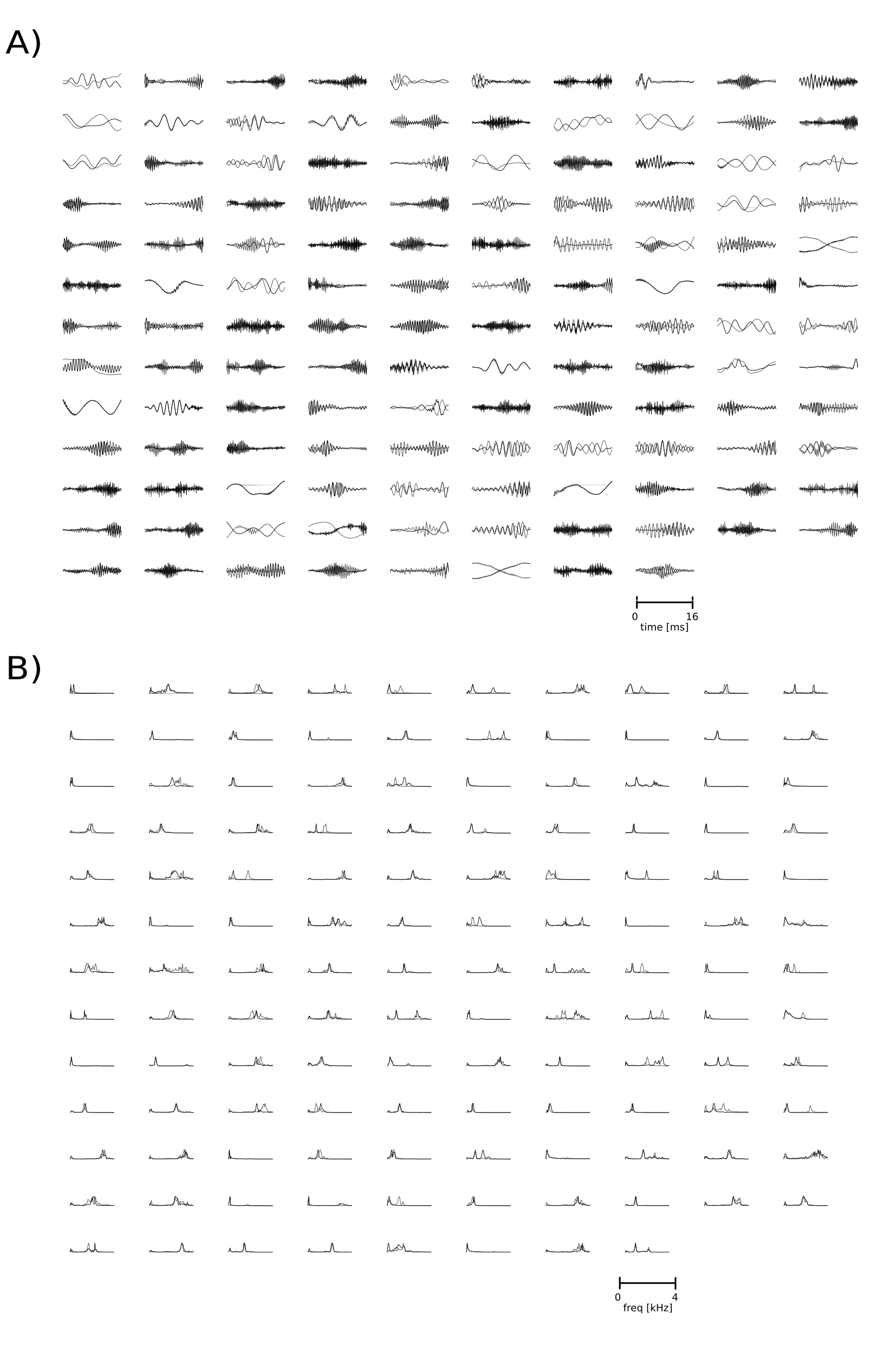}
\end{center}
\caption{Two times overcomplete set of complex, unconstrained basis functions. Black lines depict real parts, and gray - imaginary ones. A) Basis functions in temporal domain B) Basis function in frequency domain}
\end{figure}

\begin{figure}[h]
\begin{center}
\includegraphics[height=0.85\textheight]{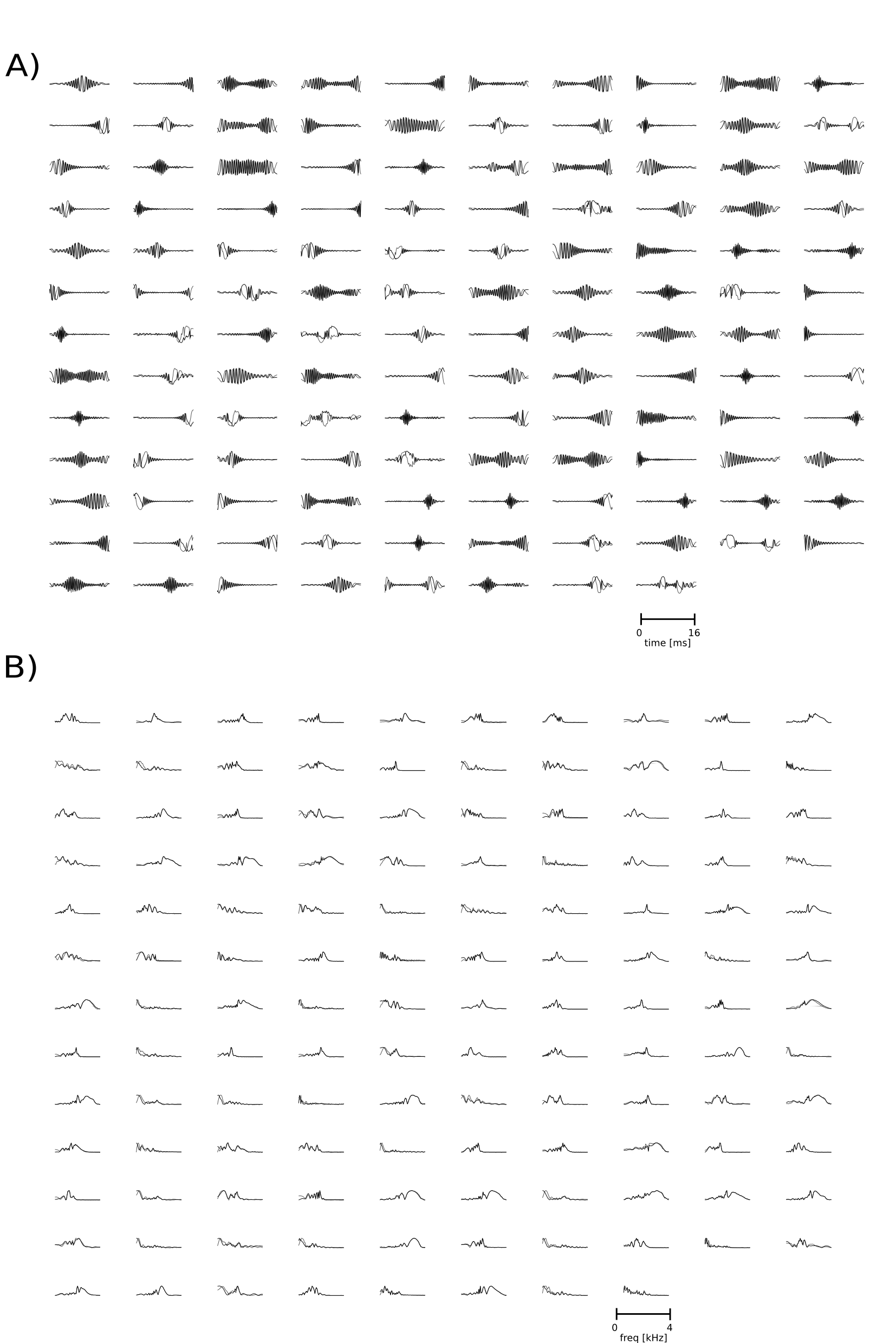}
\end{center}
\caption{Two times overcomplete set of complex, basis functions learned with priors. Black lines depict real parts, and gray - imaginary ones. A) Basis functions in temporal domain B) Basis function in frequency domain}
\end{figure}

\end{document}